\pdfoutput=1

\documentclass[11pt,a4paper]{article}
\usepackage{times,latexsym}
\usepackage{url}
\usepackage[T1]{fontenc}

\usepackage[acceptedWithA]{tacl}

\usepackage{xspace,mfirstuc,tabulary}

\newif\iftaclinstructions
\taclinstructionsfalse %
\iftaclinstructions
\renewcommand{\confidential}{}
\renewcommand{\anonsubtext}{(No author info supplied here, for consistency with
TACL-submission anonymization requirements)}
\newcommand{\instr}
\fi

\iftaclpubformat %

\else

\fi

\usepackage{amsmath,amsfonts,amssymb}
\usepackage{times}
\usepackage{latexsym}
\usepackage{booktabs}
\usepackage{multirow}
\usepackage{array}
\usepackage{float}
\usepackage{graphicx}
\usepackage{subcaption}
\usepackage{svg}
\usepackage[normalem]{ulem}
\usepackage{bbm}
\usepackage{xspace}
\usepackage{setspace}
\usepackage{boxedminipage}
\usepackage[ruled,vlined]{algorithm2e}
\usepackage{inconsolata}
\usepackage{hyperref}
\usepackage{multicol}
\usepackage{listings}
\usepackage{needspace}
\usepackage{enumitem}
\usepackage{alltt}
\usepackage[most]{tcolorbox}

\usepackage[capitalise,noabbrev,nameinlink]{cleveref}

\crefname{page}{page}{pages}
\crefname{footnote}{footnote}{footnotes}   %
\crefname{equation}{equation}{equations}   %
\crefname{corollary}{Corollary}{Corollaries}  %
\crefname{line}{line}{lines}               %
\crefrangeformat{line}{lines #3#1#4--#5#2#6}
\crefname{lstlsting}{Listing}{Listings}   
\crefname{section}{\S}{\S\S}
\Crefname{section}{\S}{\S\S}    %
\crefformat{section}{#2\S#1#3}  %
\Crefformat{section}{#2\S#1#3}
\crefrangeformat{section}{\S\S#3#1#4--#5#2#6}
\Crefrangeformat{section}{\S\S#3#1#4--#5#2#6}
\crefmultiformat{section}{\S#2#1#3}{ and~\S#2#1#3}{, \S#2#1#3}{ and~\S#2#1#3}
\Crefmultiformat{section}{\S#2#1#3}{ and~\S#2#1#3}{, \S#2#1#3}{ and~\S#2#1#3}
\crefrangemultiformat{section}{\S\S#3#1#4--#5#2#6}{ and~\S\S#3#1#4--#5#2#6}{, \S\S#3#1#4--#5#2#6}{ and~\S\S#3#1#4--#5#2#6}
\Crefrangemultiformat{section}{\S\S#3#1#4--#5#2#6}{ and~\S\S#3#1#4--#5#2#6}{, \S\S#3#1#4--#5#2#6}{ and~\S\S#3#1#4--#5#2#6}

\usepackage[T1]{fontenc}

\usepackage[utf8]{inputenc}

\usepackage{microtype}

\newcommand{\problemname}{decision-oriented dialogue}
\newcommand{\problemabbr}{DoD}
\newcommand{\problemnames}{decision-oriented dialogues}

\makeatletter
\newcommand{\smartperiod}{\@ifnextchar.{}{.\@\xspace}}
\newcommand{\smartcomma}{\@ifnextchar.{}{,}\xspace}
\makeatother
\newcommand{\latin}[1]{#1}
\newcommand{\eg}{\latin{e.g.}\smartcomma}
\newcommand{\ie}{\latin{i.e.}\smartcomma}
\newcommand{\vs}{\latin{vs}\smartperiod}

\tcbset{
  aibox/.style={
    width=474.18663pt,
    top=10pt,
    colback=white,
    colframe=black,
    colbacktitle=black,
    enhanced,
    center,
    attach boxed title to top left={yshift=-0.1in,xshift=0.15in},
    boxed title style={boxrule=0pt,colframe=white,},
  }
}
\newtcolorbox{AIbox}[2][]{aibox,title=#2,#1}

\usepackage[]{todonotes}  %

\title{Decision-Oriented Dialogue for Human--AI Collaboration}

\usepackage{fdsymbol}
\author{Jessy Lin$^{{*}1\;}$ \qquad Nicholas Tomlin$^{{*}1\;}$ \qquad Jacob Andreas$^{2\;}$ \qquad Jason Eisner$^{2\;}$ \\
  $^{1}$ UC Berkeley \quad $^{2}$ Microsoft Semantic Machines \\
  \normalsize \texttt{\{jessy\_lin, nicholas\_tomlin\}@berkeley.edu} \\\normalsize \texttt{\{jaandrea, jason.eisner\}@microsoft.com}}

\newcommand\blfootnote[1]{%
\begingroup
\renewcommand\thefootnote{}\footnote{#1}%
\addtocounter{footnote}{-1}%
\endgroup
}

\begin{document}
\maketitle
\blfootnote{\hspace{-0.1cm}$^\text{*}$Equal contribution.}
\begin{abstract}
We describe a class of tasks called \textit{\problemnames{}}, in which AI assistants such as large language models (LMs) must collaborate with one or more humans via natural language to help them make complex decisions.
We formalize three domains in which users face everyday decisions: (1) choosing an assignment of reviewers to conference papers, (2) planning a multi-step itinerary in a city, and (3) negotiating travel plans for a group of friends.
In each of these settings, AI assistants and users have disparate abilities that they must combine to arrive at the best decision: assistants can access and process large amounts of information, while users have preferences and constraints external to the system.
For each task, we build a dialogue environment where agents receive a reward based on the quality of the final decision they reach.
We evaluate LMs in self-play and in collaboration with humans and find that they fall short compared to human assistants, achieving much lower rewards despite engaging in longer dialogues.
We highlight a number of challenges models face in decision-oriented dialogues, ranging from goal-directed behavior to reasoning and optimization, and release our environments as a testbed for future work.%

\end{abstract}

\section{Introduction}

\begin{figure*}[ht]
\includegraphics[width=\textwidth]{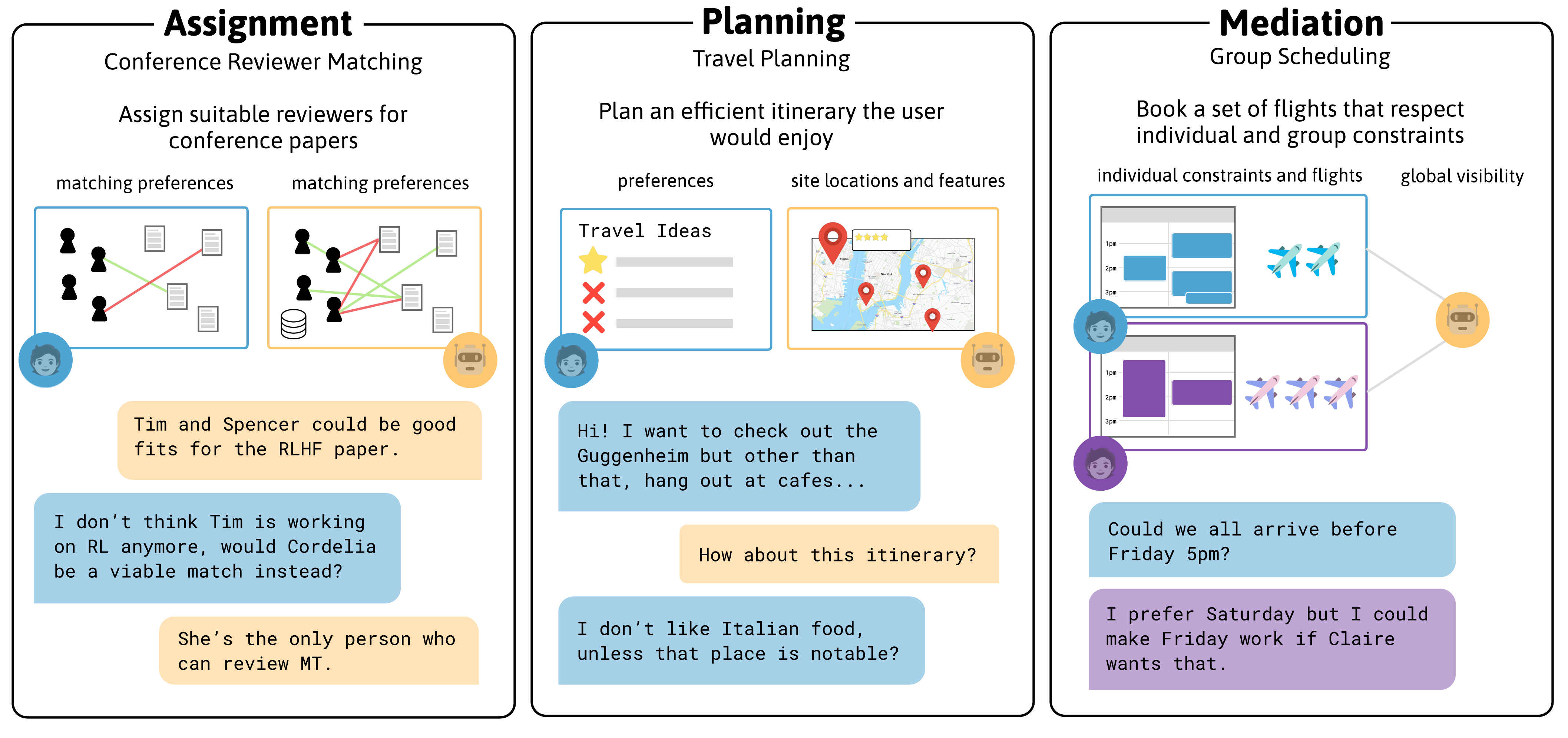}	
\centering
\caption{Overview of the three collaborative dialogue tasks that we consider. In \texttt{Assignment}, two agents with symmetric access to information play the role of area co-chairs assigning reviewers to conference papers. In \texttt{Planning}, an assistant collaborates with a user to help them plan an itinerary. In \texttt{Mediation}, an assistant must chat with multiple separate users to help them resolve a group scheduling problem.}
\label{fig:tasks}
\end{figure*}

Imagine that you are trying to book conference travel with the help of a digital assistant. Your choice of airline is flexible, but you'd rather avoid layovers, want to arrive a day or two before the conference begins, and would like to be able to check in to your hotel as soon as you arrive. Additionally, you're in charge of booking travel for a few of your colleagues, each of whom has their own preferences and budgets, some of whom will be flying in from different cities, but all of whom would like to arrive at roughly the same time and stay in a nearby area. Suddenly, you must manage and communicate about a combinatorial explosion of possible itineraries.  %

Similar optimization problems occur in many everyday situations. %
Consider consulting a friend about what computer they'd recommend with the best tradeoff of features for your use cases.
Or trying to allocate funding from multiple grants to determine which students should work on which projects, while juggling student preferences.
Or making strategic decisions with your colleagues about which projects your company will take on and who to hire to manage those projects.
All these situations share an underlying decision problem in the face of uncertainty, where collaborating with others is often critical to arrive at the best solution.

Difficult decision problems like these are precisely where AI assistants could shine.
Automated systems can handle large amounts of information and complex computations much better than humans.
For example, in cases like travel booking, they can quickly search over a large number of possible itineraries and compute total costs in a way that the average user cannot.
They may also be able to efficiently reason under uncertainty about the expected value of decision-relevant information, helping them determine what information may be important to share with or request from the user.
On the other hand, these decisions cannot be \emph{fully} automated either. AI assistants \emph{complement} humans' knowledge and capabilities: people know their preferences and may have other knowledge external to the system, including knowledge about fuzzy real-world constraints that are difficult to formalize in a computer-readable format.
To solve these problems, systems need to communicate with users, ideally with a flexible interface such as natural language.
However, there is limited existing work evaluating model performance in these types of conversational settings.
In this paper, we develop a challenging suite of decision problems in which multiple agents must collaborate with each other and make decisions via natural language.
We then benchmark the abilities of language models on these tasks and release datasets and environments to encourage future modeling work in this area.

We begin by formalizing the setting of \textit{\problemname{}},
a class of tasks in which 
multiple agents must communicate in order to
arrive at a joint decision, perhaps from a combinatorially large space of options.  Agents in these tasks are jointly rewarded according to the quality of the decision. 
Each agent starts out with different information: for example, the user knows their own travel preferences, while the AI assistant has a database of flight and hotel prices.  Sharing their information allows them to better assess different travel plans. 
Critically, however, the large amount of information 
makes it unnatural and inefficient for assistants to communicate \emph{all} of their knowledge to users, or vice versa.
Instead, agents must determine what their partners already know and what information is likely to be decision-relevant, asking clarification questions and making inferences as needed.

Within this class of tasks, we present three everyday domains where humans and agents must collaborate in order to make complicated decisions.
(1)~In \texttt{Assignment}, two agents take on the role of conference area chairs, assigning reviewers to conference papers when each agent has only has partial information about reviewer--paper fit. (2)~In \texttt{Planning}, an assistant with knowledge of a city must assist a human with building an itinerary based on their preferences. (3)~In \texttt{Mediation}, multiple users must collaborate with an assistant in order to resolve group scheduling challenges.
For each task, we specify an objective measure of utility based on the quality of the final decision. We first collect human--human dialogues on these tasks in order to establish a reference point for how humans naturally collaborate with each other.  These are long dialogues, averaging 13 messages over 8 minutes (\cref{tab:dataset-stats}).  We then develop extensible environments for evaluating language models on each task. %

We use these environments to benchmark the relative performance of GPT-3 \citep{NEURIPS2020_1457c0d6} in collaboration with humans, along with additional experiments in self-play and in a novel evaluation procedure known as \textit{prompted self-play}, in which AI agents complete partial human dialogues.
We then identify several common failure modes of GPT-3 and provide analyses of self-play dialogues.
We release all dialogues, environments, and interfaces for human data collection in order to encourage future work on these challenges.\footnote{\url{https://github.com/jlin816/dialop}}

\section{Task Formulation}
We formalize a \emph{\problemname{}} (\problemabbr{}) task as a multi-agent problem consisting of a set of agents, an underlying world state $W$, each agent's partial and possibly noisy observation $O_i$, a set of legal messages $m \in \mathcal{M}$ (analogous to actions in an Markov decision process), a reward function $R$ with parameters $\theta$ that evaluates decisions, and a communication cost function $C$.
The goal of a \problemname{} is to find a decision that maximizes $R$ while minimizing the communication cost function $C$.
$W$ remains fixed throughout the dialogue.
Our problem can be thought of as a decentralized partially observable Markov decision process (Dec-POMDP;~\citealp{bernstein2000complexity}) in which actions are messages and formal decisions.

An agent $i$'s policy $\pi_i$ maps its known information $O_i$ and the dialogue history $\{ m_1, \ldots m_{t-1} \}$ to a new message $m_t$: $\pi_i(m_t \mid O_i, \{ m_1, \ldots m_{t-1} \})$.
Agents send messages by sampling from their policy.
Messages may specify a recipient if the number of agents $> 2$, and are expressed in natural language except for three special formal messages: a proposed decision, a formal acceptance of a decision, and a formal rejection. If an agent sends a proposed decision message and all other agents respond with formal acceptances, the dialogue ends.

To illustrate the information in a \problemabbr{}, consider the task of planning a travel itinerary that satisfies a user's preferences (\texttt{Planning}, as shown in~\cref{fig:tasks}, middle).
We represent the underlying world state as a weighted graph $W = (V,E,w)$ whose vertices are potential destinations.  A decision is a path $W'$ in $W$, representing the itinerary.  Higher-weighted paths are better and the agents must communicate to improve their knowledge of the edge weights.

In general, we represent the world state $W$ as a weighted graph and the possible decisions as subgraphs $W'$ that satisfy task-specific constraints.\footnote{Representing $W$ as a graph lets us model most discrete optimization problems. A more general formulation could assume an unstructured world state; agents would communicate about random variables representing unknown quantities in the world state, rather than features of an underlying graph.}
Edges and vertices in $W$ have weights $w(e_{ij}), w(v_i)$ that represent rewards (which may be negative) for including them in $W'$. The optimal decision for this world state is a subgraph $W' \subseteq W$ that maximizes the reward
\begin{align}
R_{\theta}(W') & = \sum_{v \in W'} w(v) + \sum_{e \in W'} w(e) \label{eq:obj}
\end{align}
In principle, the reward function could be any function of $W'$, but we focus on the linear objective \labelcref{eq:obj}.  For most practical tasks, the constrained optimization problem could then be expressed as an integer linear programming problem and solved using standard algorithms.
We assume edge and vertex weights are determined by their features, represented by feature vectors $\phi(\cdot) \in \mathbb{R}^k$, so that:
\begin{equation}
\begin{aligned}
w(v_i) &= \theta^T \phi(v_i) 
& \ \ 
w(e_{ij}) &= \theta^T \phi(e_{ij})
\end{aligned}
\end{equation}
where $\theta$ is a preference vector.\footnote{To reward edges between similar or dissimilar vertices, one could define $\phi(e_{ij}) = \phi(v_i) \odot \phi(v_j)$, for example.}

The 
hard constraints on $W'$ and the form of the objective are treated as common knowledge. However, the world state $W$---in particular the feature vectors and 
the preferences $\theta$---is only partially observed by each agent. Therefore, crucially, agents must exchange messages in order to reduce their respective uncertainties about the optimization problem. However, there is a cost to communicating (\eg time or effort), which agents must trade off with their desire to achieve a good decision. Thus, the overall objective function for a \problemabbr{} is:
\begin{align}
    \max_{W', \mathbf{m}}\mbox{} & R_{\theta}(W') - \sum_t C(m_t) \\
\text{subject to}& \textit{ task-specific constraints on } W' \subseteq W\nonumber
\end{align}

Other collaborative or task-oriented dialogue tasks are typically evaluated on coarse metrics such as success rate \citep{li2016deep}, which measure whether a system accomplished its user's goal. In contrast, the reward in a \problemabbr{} provides a \emph{graded} measure of communication success, measuring how close to optimal a final decision is.

\begin{figure*}[h]
\includegraphics[width=.8\textwidth]{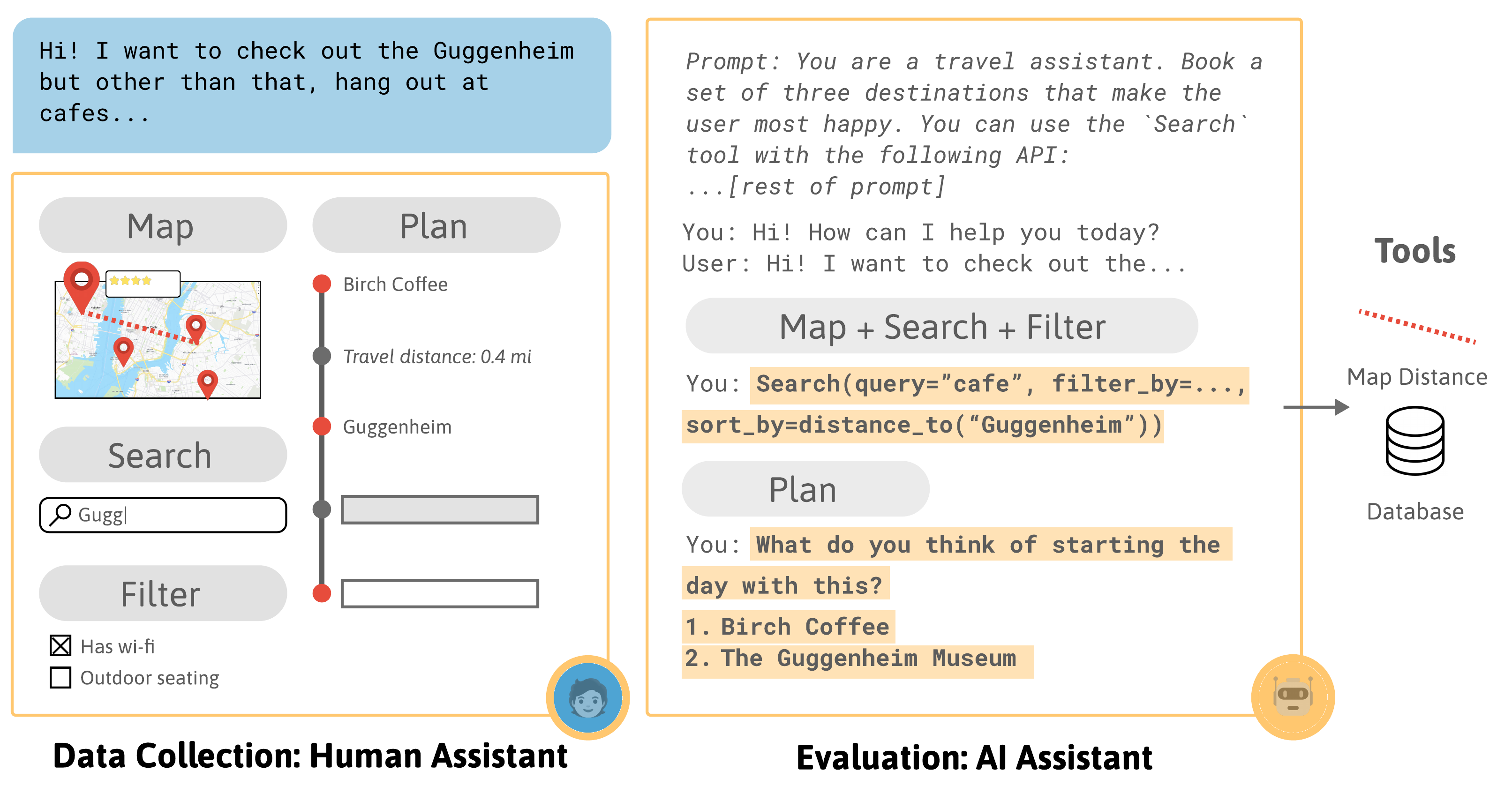}	
\centering
\caption{Data collection and evaluation frameworks. In order to collect human-human dialogues, we built web interfaces that allow humans to play either the User or Assistant role for each task. When evaluating how well an AI language model plays one of these roles, we linearize information from the web interface into a text prompt and provide additional tools that let the language model access information that cannot fit within its context window.  This figure shows just the Assistant role, for one task.}
\label{fig:data-collection-eval}
\end{figure*}

\section{The \texttt{DialOp} Environments}

We introduce three everyday collaborative decision-making domains formalized as \problemabbr{} tasks.
To instantiate them, we release \texttt{DialOp}, an open-source suite of \problemname{} environments. %
For each task, we implement a graphical UI to build human user interfaces for data collection (as in~\cref{sec:data-collection}), a text environment to evaluate models in self-play (as in \cref{sec:exp-sp}), and a unified interface between the two to evaluate models in collaboration with humans (as in \cref{sec:exp-human-lm}). Here, we describe how we formalize each everyday scenario as a \problemabbr{} problem and implement the environments.

In contrast to other dialogue tasks where evaluation is based on supervised datasets, we procedurally generate each game by sampling the parameters of the underlying decision problem (e.g. the reward parameters $\theta$) to instantiate new dialogue contexts\footnote{We will use \emph{task} to mean the formal problem setting;  \emph{environment}, our code implementation of a task; and \emph{game}, a generated episode or instance with specific parameter settings.}.
To account for the variance in the difficulty of randomized optimization instances (i.e. for ease of comparison and optimization in future modeling approaches), we normalize rewards to $[0,1]$.
This generation process enables future work to study how models generalize: e.g. to larger optimization problems (by changing the parameter dimensions) or new domains (by changing the ``theme'' while keeping the underlying parameters fixed). We provide more details on environment generation in~\cref{sec:appendix-env-generation}.

AI agents interact with the text environments through an OpenAI Gym-like interface~\citep{brockman2016openai}, which is designed to provide text-only language models like GPT-3 with the same affordances that humans have in the GUI. Agents send messages to the environment,
prefixing each with a message type (\texttt{[message]}, \texttt{[propose]}, \texttt{[accept]}, or \texttt{[reject]}), which the environment parses to determine how to interpret the message. Messages are forwarded to other agents. Proposals can be partial (e.g. a subset of the itinerary) or full, and may optionally be accompanied by another message such as a clarifying question.
Proposals are parsed and scored; if full, the only valid actions for the other agents are \texttt{[accept]} and \texttt{[reject]}.
Formal rejections clear the current proposal, and formal acceptances terminate the game. Below, we describe how the environments implement each of the decision domains we introduce.

\subsection{\texttt{Assignment}}
Our first task is an idealized bipartite matching problem, motivated by the scenario of conference organizers assigning reviewers to submitted papers (\cref{fig:tasks}, left). Although reviewer matching is sometimes automated via approaches like the Toronto Paper Matching System \citep[TPMS;][]{Charlin2013TheTP}, human organizers often have their own incomplete and partially-overlapping knowledge about which reviewers fit which papers. Fit cannot necessarily be described on an absolute scale, so when working together on an assignment, organizers must discuss relative edge weights (``Alice would be a better choice than Bob for paper 8'').  TPMS could in principle be replaced by an AI agent that joins this dialogue as an additional participant. We consider a simplified version of this problem in which two agents must find a one-to-one matching between reviewers and papers.

\paragraph{Formalization}

We represent $W$ as a bipartite graph and restrict valid proposals $W' \subseteq W$ to be bipartite matchings. Edge weights $w(e_{ij})$ represent reviewer-paper affinities, and each agent observes some subset of these weights. Agents have symmetric information and roles in this task: their observations are drawn from the same distribution, and either agent can propose a decision.\footnote{There are many ways we could have made the task more realistic.  
Each score could be a function of underlying features, for example, the dot product of the paper's topic vector and the reviewer's topical-expertise vector. Each agent could then observe and discuss a subset of these features---``Alice is an expert on Botany''---rather than observing full edge weights.  Orthogonally, we could use noisy observations.  Features of the agents themselves might affect what they tend to observe.
}

\paragraph{Environment Implementation}
For each game, we sample a random $8\times8$ table of reviewer-paper affinity scores (edge weights). Each cell is shown to each agent with probability $p_{\text{observed}}=0.4$, so that a given cell may be shown to just one agent, to both, or to neither.

To discourage reviewers from communicating affinity scores in the form of numbers---which would not be natural in the real-world version of this scenario---we scale all scores shown to each agent by a random positive constant, so that they are not comparable across agents but can still be discussed in relative terms such as ``X is much better than Y.'' Each agent observes a subset of the reviewer-paper affinity scores, scaled by some constant unknown to them.
The agents' shared reward is the value (sum of edge weights) of the final matching, normalized by the value of the best matching with the agents' \emph{pooled} knowledge. More precisely, we compute the best matching by taking each edge's weight to be its posterior mean weight given all observations of both agents.

\begin{figure*}[h]
\includegraphics[width=\textwidth]{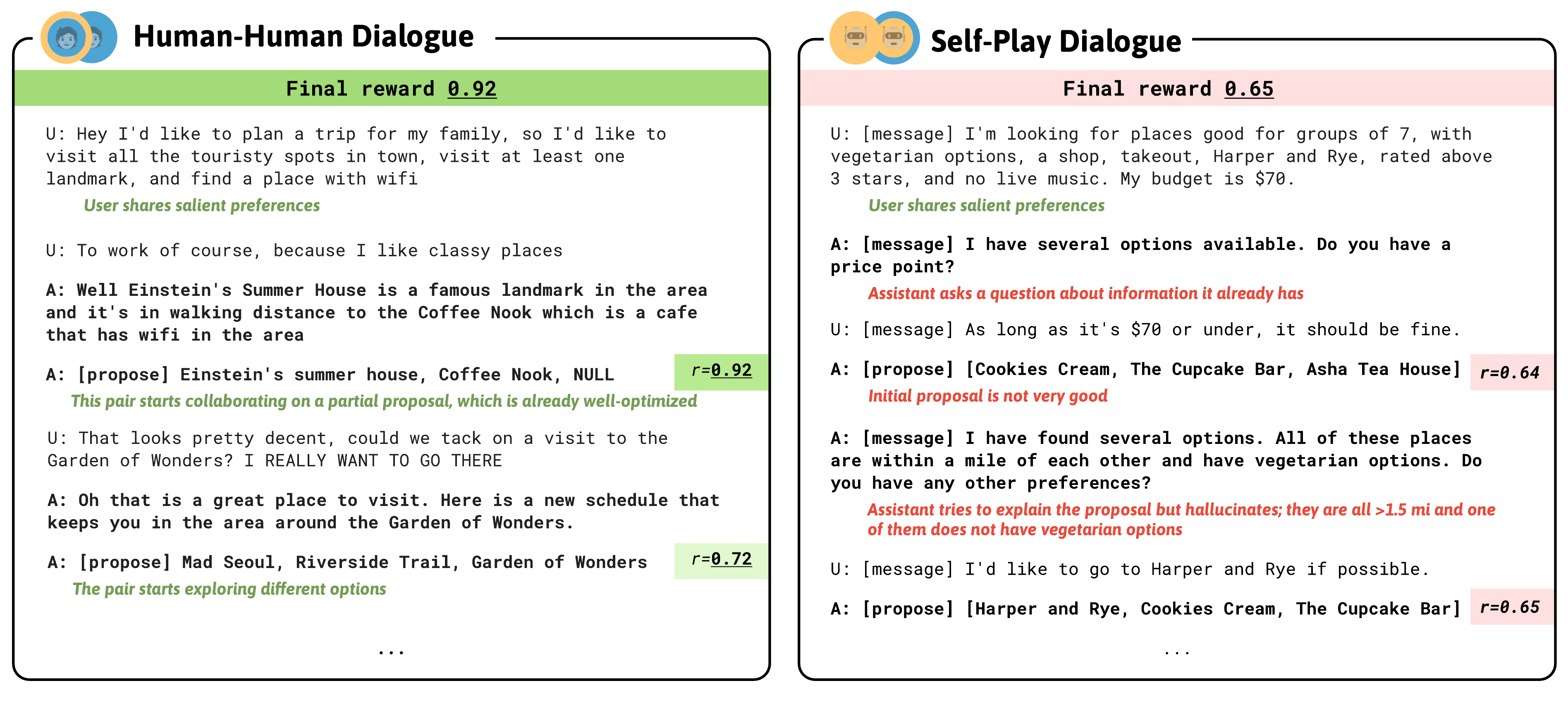}	
\centering
\caption{For the \texttt{Planning} task, an annotated example of a human-human dialogue (left) and an annotated example of an LM self-play dialogue using GPT-3 (right). While humans generally exhibit diverse and flexible strategies and reach good solutions, self-play dialogues tend to be repetitive, and the assistant makes mediocre proposals and often hallucinates. We discuss further in~\cref{sec:analysis}.}
\label{fig:qualitative}
\end{figure*}

\subsection{\texttt{Planning}}
Next, we consider a scenario in which a user is planning an itinerary in a city with the assistance of a travel agent (\cref{fig:tasks}, middle).
While existing systems can assist with parts of travel such as recommendation or booking, they often expect users to provide close-to-full specifications of their requests, rather than working toward a solution together.
Ideally, systems would be able to assist us in the comprehensive way that a human travel agent would: start with an under-specified set of desiderata,
propose possible multi-day itineraries based on partial knowledge of the user's preferences and domain knowledge, and iteratively refine the plan with the user, filling in and revising details based on feedback. We consider a small version of this problem where the assistant must help the user plan an itinerary of several sites.

\paragraph{Formalization}
We formalize this task by constructing $W$ as a fully-connected graph over the sites, where edge weights represent travel times.
The user has preferences $\theta$ about which sites to visit, a financial budget, and a preference for reducing travel time (i.e., a negative preference on edge weights). Meanwhile, the assistant has access to a database of sites, along with information about their cost, location, and amenities (\eg outdoor seating). Unlike reviewer matching, this task exhibits asymmetry of information: the assistant has information about vertex features and edge weights, while the user only has information about their own preference vector $\theta$. Additionally, only the assistant can make proposals, which the user must accept or reject. Due to the budget constraint, the prescribed itinerary length $k$, and the preference to minimize travel, this domain involves aspects of the knapsack problem, subset-selection problems, and the traveling salesperson problem.

\paragraph{Environment Implementation}
In each game, the assistant must propose a set of three sites.
The environment comes with a set of sites (\eg restaurants, parks, museums). On each game, the environment randomizes the features of each site (\eg expected price range). The environment also has a set of preference features with natural language labels (\eg a preference for ``Wi-Fi available'') and randomly generates the user's preference vector $\theta$ with $s=10$ nonzero elements.

To simulate the fact that people cannot quantify their actual preferences on an absolute scale, the user only observes natural language descriptions of their nonzero preferences with binned magnitudes (strong negative, mild negative, mild positive, strong positive).
The assistant only observes the inventory of sites and their features. The environment optionally provides API calls to search over sites, either via (1) a simple domain-specific language (DSL) that can query specific \texttt{fields} (e.g. name, category, price) of a site, \texttt{filter} over fields, \texttt{sort\_by} field values (including \texttt{distance\_to} another destination), and search by \texttt{text\_query} in freeform natural language or (2) an LM prompted with examples in the DSL as query executor, which permits simple generalizations from our DSL.

When the assistant proposes a complete or partial itinerary, the proposal reward (while unknown to the assistant) is automatically computed for the user's convenience, including a breakdown of the contributions to the reward from each site, travel times, and budget constraints. Showing scored proposals to the user simulates that real users intuitively know how they feel about an itinerary, even if they may not be able to name their preferences up front.
With this information, the user can make judgments about aspects of the itinerary (\eg that it is worth spending extra travel time to visit a particularly desirable site).
The game ends when the user accepts a full itinerary of $k$ sites.
The agents' shared reward is the score of the itinerary, range-normalized by the scores of the best and worst possible $k$-site itineraries.

\subsection{\texttt{Mediation}}

Finally, we introduce a coordination scenario where the assistant plays the role of mediator among multiple users (\cref{fig:tasks}, right). The users are attempting to book flights from their respective cities to all arrive at some shared destination at around the same time, \eg to meet up for an event or vacation.
Assistants could be helpful not just for maximizing individual preferences, but for efficiently considering configurations for the entire group.
We consider a setting where $n$ users can only coordinate through the single assistant.
In the task, each user wants to choose a flight that is inexpensive and avoids conflicts with the user's calendar commitments, but that arrives close to the arrival times of other users.
The assistant has access to each user's flight options and work calendar, but doesn't observe the user's personal calendar, nor the user's preferences about which meetings are most important.

\paragraph{Formalization} In the underlying optimization problem, the world state $W$ can be modeled as a complete $n$-partite graph, where the vertices associated with each user are their flight options. 
Any two flights for different users are connected by an edge, whose weight indicates how compatible the flights are (\ie whether they arrive at similar times).  Vertex weights are derived from the users' calendars, with more important meetings creating a larger preference against flights (vertices) that conflict with them.  The goal is to select a flight for each user so that the induced subgraph $W'$ (with $n$ vertices and $n \choose 2$ edges) has high total weight.  This task has asymmetric roles and information.

\paragraph{Environment Implementation}
In each game, the assistant must coordinate flights for $n=2$ users.
The environment generates a random set of personal calendar and work calendar events, as well as weights for each event indicating how important it is.
The environment also generates a list of flights for each user, each with randomized features for price, arrival time, and departure time.

The user observes their own personal and work calendar and flight set, while the assistant observes the work calendars and flight sets of \emph{both} users (but not their personal calendars, and without the meeting importances). The assistant has one-on-one chats with each user and is allowed to talk to any user at any time; deciding which user to talk to is itself a strategic decision.

The assistant can make a partial proposal to a single user or a full proposal that warrants a formal decision on the next turn to both users jointly. 
Each user who receives the proposal is shown the score for their own flight, broken down in terms of price and missed meetings, as well the closeness to the other user's flight in the case of a joint proposal. The game ends when both users accept some joint proposal. The final reward is the total weight of the proposal (i.e., $R_{\theta}(W')=w(v_i)+w(e_{ij})+w(v_j)$), range-normalized by the total weights of the best and worst possible proposals.

\section{Dataset}
\label{sec:data-collection}

In order to study the communication strategies used by humans and establish baseline performance numbers, we collected a set of human-human dialogues.
For each task, we built a multi-player online interface (\cref{fig:data-collection-eval}, left) and collected high-quality human-human dialogues in randomized games using a mixture of workers hired directly and through Amazon Mechanical Turk, resulting in a total of 409 dialogues, consisting of 5253 messages and over 58K words across domains.
Pairs of human players take a median time of 8min~19sec across tasks, showing that these tasks are nontrivial. They achieve an average of roughly 90\% of the maximum possible range-normalized reward on both the assignment and planning domains, and close to 100\% performance in the mediation domain.
We provide additional data statistics and example dialogues for each task in \cref{sec:appendix-data-collection}.

In each task, each worker played the role of an assistant or user. For ease of play, players were not required to take turns, but used a chat interface where they could send a message at any time.
Consecutive messages from the same player were then concatenated into a ``turn.''

Real-world users would know their own preferences, but our workers are emulating users that we have generated programmatically, so we must tell them what their preferences are.  This setup gives us full knowledge of user preferences so that we can objectively evaluate the quality of the decision.

\section{Baseline Models}
\label{sec:baselines}
\begin{figure*}[t]
    \centering
    \includegraphics[width=0.95\textwidth]{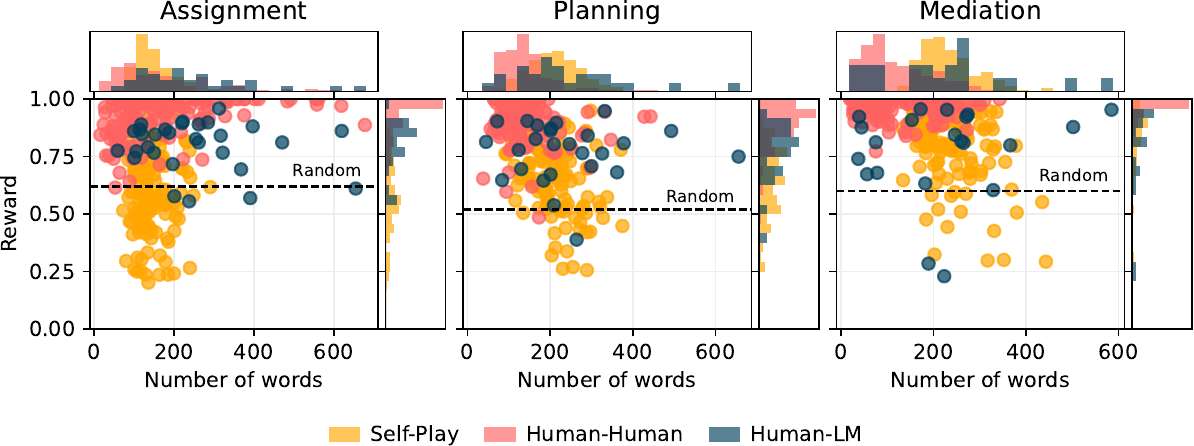}
    \caption{Human-LM and self-play scores compared to human dialogues, plotted against dialogue lengths in words. LM assistants achieve lower scores than human assistants on average, and also tend to have longer dialogues. Models in self-play have even lower scores and longer dialogues since they must also play the role of a cooperative user.
    The histograms show the marginal distributions of the scores and dialogue lengths. The dashed line shows the average score of a random proposal.}
    \label{fig:selfplay-scatter}
\end{figure*}

Future AI agents for decision-oriented dialogue may benefit from incorporating explicit reasoning over possible world states and possible decisions.  However, as a baseline approach, this paper evaluates few-shot prompted LMs as the AI agents.  These have the benefit that they can attempt a wide variety of dialogue interactions without the need for domain-specific training or modeling. We focus our evaluations on the instruction-tuned \mbox{GPT-3} model known as \texttt{text-davinci-003} \citep{NEURIPS2020_1457c0d6, ouyang2022training}, prompted for each task with 1--2 of the human-human dialogue examples that we collected for that task.
LMs have access to the same information and actions that human annotators do, presented through formatted text strings (\cref{fig:data-collection-eval}, right) rather than through the graphical UI used by human annotators  (\cref{fig:data-collection-eval}, left).

If a model generates an invalid message (\eg{} if the user in \texttt{Planning} or \texttt{Mediation} sends a proposal), we append the message to the prompt, along with any error message from the game, and continue generating, allowing the model to revise its previous generation.
Generally, we simply prompt models with player information in context, with some exceptions we note here.
For \texttt{Planning}, we noted that models needed particularly complex reasoning to search based on the dialogue (on the assistant side) and to decide whether to accept an itinerary based on the scores (on the user side), so we implemented a ReAct-style prompting approach~\citep{yao2022react}. To do so, we augment the few-shot example dialogues in the user and assistant prompts with \texttt{[think]} steps (``\texttt{[think] I am losing the most points from the travel time between events. I should reject the proposal...}''), which demonstrate how the agent can reason.
For \texttt{Mediation}, to handle the multi-party dialogue, we adopt a simple turn-taking strategy where we iterate round-robin through all agents; on the assistant's turn, it is prompted with ``\texttt{You to}'' and chooses which user to send the message to by generating either \texttt{0} or \texttt{1}.

\section{Evaluation}
\label{sec:exp}
In this section, we evaluate the baseline models to determine how well prompted present-day LMs can collaborate with humans.
First, we directly compare the performance of LM assistants with human assistants at assisting human users.
Second, although helping actual humans is the ultimate goal, human-LM evaluation is expensive and frustrating for human users, given the quality of current models, so we add two automatic evaluation settings for our benchmark to ease future evaluation and provide additional insights into model behavior: self-play and prompted self-play.

\subsection{Human-LM Evaluation}
\label{sec:exp-human-lm}

First, we evaluate whether current baseline prompted LMs can serve as effective decision-making assistants. We recruited 13 participants (a mixture of undergraduates, graduate students, and contractors) and collected a total of 77 dialogues between these participants and GPT-3, prompted with the information for the assistant role.
In~\cref{fig:selfplay-scatter}, we show human-human and human-LM normalized rewards against the number of words in the dialogue.
We also show the performance of a naive rule-based baseline that selects a random proposal from the set of all possible proposals.

We observed that human-LM dialogues achieved lower scores, despite being longer than human-human dialogues.
Qualitatively, participants had a frustrating experience with the LM assistant.
In initial trials, we observed that the LM assistant would often get ``stuck'' making similar proposals repeatedly, leading the dialogue to fail to make progress. In these cases, users were instructed to accept the best proposal they could get, but dialogues likely could have been much longer.
We discuss particular failure modes of LM assistants further in~\cref{sec:analysis}.
Overall, these results suggest that present-day LMs are far from serving as useful assistants, despite the appearance of helpfulness.

\subsection{Self-Play}
\label{sec:exp-sp}

Since human evaluation is expensive and frustrating, we evaluate whether models can collaborate with each other in self-play, prompting another model to play the role of the user as a cheaper proxy for humans.
We prompt models with the same randomly generated task instances as the human-human dialogues in the evaluation dataset to reduce variance, although future agents can also generally be evaluated on new random instances generated from the environment.
In~\cref{fig:selfplay-scatter}, we see that models in LM self-play achieve lower rewards and produce longer dialogues than both human-human and human-LM pairs.
We note that self-play is a more difficult setting than human-LM play, as models also have to serve as cooperative \emph{users}.
The performance drop compared to human-LM pairs suggests that human partners may somewhat compensate for model failures, e.g., by taking initiative to share relevant information or keeping the dialogue on track to better solutions.

\subsection{Prompted Self-Play}
\label{sec:exp-psp}
\begin{figure*}[t]
    \centering
    \includegraphics[width=\textwidth]{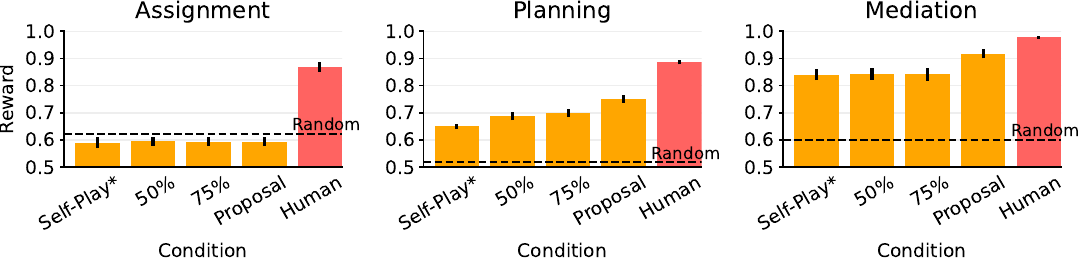}
    \caption{Prompted self-play results for all three tasks, compared to human results. For each setting, we initialize dialogues with 50\% and 75\% of a corresponding human game and let GPT-3 complete the dialogue. In the \textit{proposal} setting, we prompt the model with an entire human dialogue except for the final proposal and force the model to end the game immediately. The average score of a randomly selected proposal is shown for each task as a dashed line. (*) For reference, we also show the mean score of models in unrestricted self-play; this differs from a 0\% PSP condition, because PSP biases the models to stop when the dialogue reaches the corresponding human-human dialogue length.}
    \label{fig:psp}
\end{figure*}

As a more nuanced proxy for human evaluation, we also propose a new mode of automatic evaluation, \textit{prompted self-play} (PSP), in which a given prefix of a human-human dialogue is completed with model-model play.
PSP provides a more fine-grained picture of model capabilities by providing models with a human dialogue that is already ``on-track,'' containing information that the human-human pair has talked about already.
This makes it easier to find good solutions \emph{if} models are able to understand and reason over that information to make a proposal.
Additionally, to decide how to proceed from the prefix, models should be able to reason over what commitments were established or what information is known by the other agent.
For example, models ought to avoid asking about information already implied by previous utterances---which, in PSP, include real human utterances.
Finally, prompting in this way encourages models to complete dialogues ``in the style'' of the human-human pair in the prefix. As a result, PSP can test whether models flexibly collaborate with a diverse range of humans, perhaps adopting different collaboration styles (e.g. with one agent taking most of the initiative), similar to population play and fictitious self-play evaluation~\citep{doi:10.1126/science.aau6249, NEURIPS2021_797134c3}.

Given a human-human dialogue from our dataset, we test how models perform if they are provided with 50\% of the dialogue, 75\% of the dialogue, and everything except the final proposal, and then continue the dialogue with self-play.
We bias models to output dialogues that are approximately the same length as the corresponding human-human dialogue by prompting them to make their final proposal once the number of words in the dialogue exceeds the number of words in the human dialogue minus 25.
\cref{fig:psp} shows average PSP performance for each task. In \texttt{Planning}, models perform better with additional human data in the prompt, suggesting that they are at least partially capable of integrating information from the human-human prefix. However, there is still a substantial gap between the \textit{proposal} condition and human-human dialogue scores, indicating that models struggle to perform the final optimization step of choosing the best solution given the entire dialogue history. Meanwhile, in \texttt{Assignment}, models fail across all PSP conditions; this occurs because the final optimization step involves integrating the discussed values to compute a bipartite matching of papers to reviewers, which is difficult for models. Finally, in \texttt{Mediation}, models score well above a random baseline in all PSP conditions but do not perform better with additional human-human dialogue context, suggesting that they can meaningfully communicate about the task but don't make the optimal final proposal.
In the future, tool use could potentially greatly improve performance on this task, particularly with tools that can specifically handle the optimization part of the problem.

\section{Analysis}
\label{sec:analysis}
\begin{figure*}[t]
    \centering
    \includegraphics[width=\textwidth]{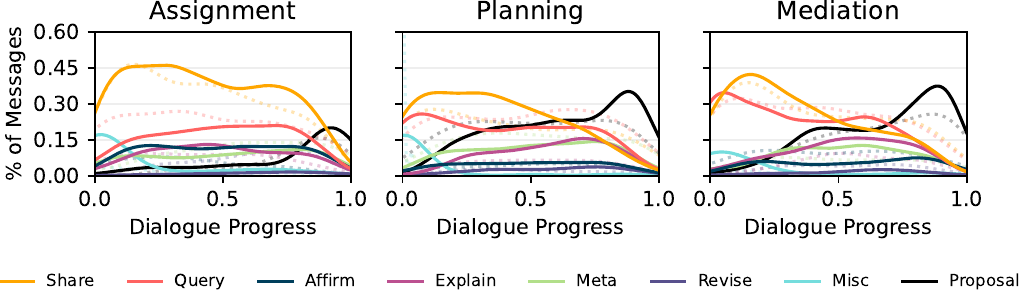}
    \caption{Kernel density estimates of message types in human-human (solid)  and human-LM (dashed) dialogues plotted against their position within a dialogue. Message types were annotated using few-shot prompting with GPT-4 and validated by manual human annotation.}
    \label{fig:kdes}
\end{figure*}

\subsection{Dialogue Act Analysis}
Humans may use a wide range of communicative strategies to negotiate with one another, optimize for their goals, and make decisions \citep{walton1995commitment}.
In order to quantify the strategies that may be useful in our tasks, we used GPT-4 to annotate human-human and human-LM dialogues at the level of individual messages. Based on manual inspection of a small set of dialogues, we devised a list of message types: (1) \textit{share}, in which agents provide information about their preferences; (2) \textit{query}, in which agents ask each other for information; (3) \textit{affirm}, in which agents agree with each other and/or conversationally ground incoming messages; (4) \textit{explain}, in which agents provide justification for a previous message or action; (5) \textit{meta}, in which agents engage in discussion about high-level strategies or meta-game details; (6) \textit{revise}, in which agents correct earlier statements; (7) \textit{miscellany}, which includes other messages such as greetings; and (8) \textit{proposal}, which denotes a formal proposed decision. These categories were roughly based on standard coarse-grained dialogue act taxonomies \citep[e.g.,][]{stolcke-etal-2000-dialogue}, which often contain statements, queries, revisions, agreements, and a miscellany category; we then added types such as \textit{meta} based on the idiosyncrasies of our problem domain.\footnote{\textit{Meta} messages reference the task but don't provide information about the underlying graph, e.g., ``I have sent a proposal'' or ``Hello! I can definitely help you find a cheap flight.'' \textit{Explain} messages justify some previous or future action, e.g., ``I think a museum would be great for the kids" after sending a proposal that includes a museum. \textit{Proposals} are task-specific formal messages, e.g., \texttt{[Mad Seoul, Riverside Trail, Garden of Wonders]} in \texttt{Planning}.}
Each message may have multiple message types. We prompted GPT-4 to generate annotations for each message using two hand-annotated example dialogues.\footnote{We performed a manual human validation on 106 messages (across six dialogues) and found that human labels matched GPT-generated labels on 88\% of messages. On the 13 instances where human labels differed, we found 7 of the GPT-generated labels to be reasonable and correct alternatives.}

We provide a breakdown of message types over the time-course of dialogues in~\cref{fig:kdes}. As expected, many interactions begin with greetings, which is evidenced by a spike in the \textit{miscellany} category at the beginning of all three plots; meanwhile, complete dialogues end in \textit{proposal} actions.
Most dialogues are focused on exchanging information: of the message types, we find that agents most commonly \textit{share} or \textit{query} for information. In the \texttt{Assignment} task, agents send twice as many \textit{share} messages as any other type of message, often sending information about individual cells in their observed tables. One common strategy involves both players sharing all observed information and then making a decision at the end of the game. This approach is most tractable in \texttt{Assignment}, where players have a relatively small observation space. However, this strategy leads to exceptionally long dialogues, even in \texttt{Assignment}, and is not the most common approach.
Meanwhile, in \texttt{Planning} and \texttt{Mediation}, which have asymmetric information and roles, agents are more likely to \textit{query} for information or engage in \textit{meta}-game discussion in order to learn what information the other agent can see. %

We observed no major differences between the types of messages used in human-human and human-LM dialogues. To investigate why human-LM dialogues fail, we turn to qualitative analysis.

\subsection{Qualitative Failures of LM Assistants}

By analyzing human-LM and self-play dialogues, we observed several classes of failure modes. Many failures are attributable to known weaknesses of LMs such as hallucinations---decision-oriented dialogues can be seen as a realistic assistance setting to elicit and evaluate these failure modes.

\paragraph{Lack of Goal-Directed Behavior}
Decision-oriented dialogues require models to explicitly optimize a decision objective. Critically, this requires \emph{planning}, e.g. asking questions that will lead to discussion of decision-relevant information, or making proposals as a mechanism for gathering information.
We observed that models do ask questions, but tend to ask general ones such as ``Do you have any other preferences?'' and sometimes slightly more specific ones such as ``Do you have a price point?'', but the questions are not \emph{goal-directed} in eliciting decision-critical information.
Models will also make iterative proposals, but the proposals only superficially build on each other (e.g. adding events one-by-one, and then concluding), often not improving in score.
This led AI assistants to be much less efficient in their dialogues (longer, yet lower-scoring) than human assistants, who in contrast, ask questions and make proposals that help them narrow down the search space.
This is unsurprising given that present-day models are not explicitly trained to optimize for task objectives beyond following the initial task instruction.

\paragraph{Failures of Reasoning}
On \texttt{Planning}, we observed that the model would make tool queries as prompted to do so, but fail to reason over the outputs of the tool (e.g., searching for museums when the user asked to visit a museum and then outputting a proposal consisting of the search results and nothing else).
Models also fail to do the optimization step of the proposal (as supported by our PSP results): proposals are often only slightly better than random, and do not improve drastically over the course of the dialogue.

\paragraph{Hallucination and Grounding}
We observed that LM assistants often failed to ground against the information they were given, outputting false information such as hallucinated flights. These instances were a major source of frustration with human users and made it very difficult to reliably collaborate with the assistant.

\paragraph{Uncooperativeness}
Human players were often frustrated that LM assistants were uncooperative. For instance, they would fail to fulfill requests like ``please add ... to the itinerary'' or would ignore information provided by the user such as ``I cannot make any flights on Friday,'' even when human players would repeatedly send these messages.
LM assistants also exhibited a failure to understand \emph{joint commitment}
by verbally committing to one course of action then making a different proposal entirely.
\texttt{Mediation} was particularly challenging due to the multi-party dialogue---here, the LM failed to manage the coordination amongst multiple players, sometimes making a proposal after eliciting preferences from one player without consulting the other player.

Beyond achieving a basic level of cooperation, we would hope that future LMs can exhibit more rich and adaptive behaviors as human pairs do. 
We show a human-human dialogue side-by-side with a self-play dialogue in~\cref{fig:qualitative}. We generally observe across the human dialogues that human-human pairs exhibit diverse strategies in (1) \emph{user \vs assistant initiative}: in some dialogues, users are proactive in sharing relevant information, while in others, assistants make directed queries to narrow down the set of proposals; and (2) \emph{coordination strategies}: working incrementally from partial proposals, backtracking, and more. In contrast, self-play dialogues and LM utterances in human-LM play tend to be repetitive.

\section{Related Work}
\label{sec:related}

\paragraph{Task-Oriented Dialogue}
Our work may be viewed as an extension of task-oriented dialogue, where a system must assist a user with accomplishing a goal, such as hotel booking or calendar scheduling~\citep{budzianowski-etal-2018-multiwoz, wei-etal-2018-airdialogue, andreas2020dataflow}.
Most task-oriented dialogue settings evaluate systems with coarse metrics such as success rate (e.g. at returning hotel information requested by a user) or word overlap with human-human dialogues.
In contrast, our tasks are grounded in underlying optimization problems, where the quality of the final solution provides a richer measure of communicative success.
Additionally, agents must \emph{take initiative} to share and query information, similar to early work on task-oriented dialogue in mixed-initiative settings \citep{novick1997mixed, horvitz1999principles} such as TRAINS~\citep{allen1995trains} and TRIPS~\citep{allen2002trips}, in which users had to collaborate with a computer agent in order to solve planning problems.

\paragraph{Grounded \& Goal-Directed Dialogue}
Many prior works have studied grounded and goal-directed dialogue more broadly, where agents use language to communicate and achieve goals, often in a setting that involves multimodal, situated, or external (non-linguistic) knowledge. Examples of such tasks include Cards \citep{potts2012goal,vogel2013emergence}, CerealBar \citep{suhr2019executing}, MutualFriends \citep{he-etal-2017-learning}, and OneCommon \citep{Udagawa_Aizawa_2019}, as well as partially-cooperative negotiation dialogue tasks such as Deal or No Deal \citep{lewis-etal-2017-deal} and Craigslist Bargaining \citep{he-etal-2018-decoupling}.
In many of these tasks, including ours, the nature of the multi-agent collaboration requires that agents not only find the optimal solution, but also reach mutual understanding (a setting termed ``grounded agreement games''; ~\citet{schlangen2019agreement}), eliciting rich coordination and communication strategies in language.
Other work has studied how agents can explicitly model user preferences to more effectively persuade or argue that a course of action is desirable~\citep{carenini2006generating}.
Decision-oriented dialogue shares elements with many of these tasks, with a focus on fully-cooperative problems in real-world decision domains and a formalism to characterize the underlying inference problem in these settings.

\paragraph{Large Language Models} Our goal of building task-general dialogue agents motivates the use of large language models (LMs) such as GPT-3 \citep{NEURIPS2020_1457c0d6, ouyang2022training}, PaLM \citep{chowdhery2022palm}, or LLaMA \citep{touvron2023llama}. 
Current-era language models are known to struggle with aspects of our tasks, such as mathematical reasoning \citep{hendrycks2021measuring}, explicit state tracking \citep{li-etal-2021-implicit}, pragmatics \citep{fried2022pragmatics}, and theory of mind \citep{sap2022neural}.
However, recent work in scratchpad prompting \citep{nye2021show}, chain-of-thought reasoning \citep{wei2022chain}, and external tool use \citep{schick2023toolformer} has sought to address these problems. We build baseline models with similar approaches in our setting. While LMs can perform reasonably well in some of our settings, we show that they cannot consistently handle dialogues with complex decision problems as well as humans.

\paragraph{Human--AI Collaboration}
Our task may also be viewed as a cooperative multi-agent setting~\citep{dafoe2020open}.
Research in human--AI collaboration and multi-agent reinforcement learning has also formalized tasks that require collaborating strategically with other agents on a shared goal, through tasks such as Overcooked \citep{carroll2019overcooked}, Hanabi \citep{bard2020hanabi}, and Diplomacy~\citep{cicero2022}.
Our evaluation methodology is adapted from these tasks, where methods like population play and fictitious self-play are often used as proxies for human evaluation in addition to self-play \citep{pmlr-v37-heinrich15, NEURIPS2021_797134c3}.
In human--AI collaboration, cooperative tasks have been formulated in game-theoretic terms where agents use signals from the user such as demonstrations, feedback, or language~\citep{jeon2020reward, lin-etal-2022-inferring} to explicitly optimize for assistive behavior~\citep{NIPS2016_c3395dd4, sadigh2016planning}.
In our work, we are similarly interested in formalizing settings where agents should explicitly optimize for effectiveness in the course of dialogue.

\section{Discussion \& Conclusion}
In this paper, we presented data, environments, and model baselines for a class of tasks we call \textit{decision-oriented dialogues}. Across all task settings, current LMs did not perform as well as humans, suggesting failures in their ability to communicate efficiently and reason in structured real-world optimization problems.
Future work in this domain may seek to integrate tools and inference techniques which would allow language models to compute optimal decisions while maintaining their flexible communication and collaboration skills.
These tasks are also useful for studying how models optimize for longer-term dialogue objectives rather than single responses. For instance, information seeking should be an emergent behavior of a model that utilizes the underlying POMDP structure of the problem to reason about how to communicate.

The ultimate goal of this line of work is to build general collaborative agents rather than agents specialized to particular settings. As we develop more generally capable models, future work should evaluate whether models can \emph{generalize} their collaborative capabilities to harder task instances and \emph{transfer} them to related tasks.
People often use strategies that depend on the visual presentation of information~\citep{kong2007global}, suggesting that multimodal agents that can use or generate visuals may improve collaboration (\eg using maps in itinerary planning).
Additionally, people often \emph{construct} their preferences over time rather than beginning with all the relevant knowledge~\citep{payne1999constructed}. Agents could help the user consider salient decision points.
Finally, we presented a particular graph-based formalism for decision-making dialogues that focuses on structured decisions and discrete optimization problems. Many real-world problems may lack this formal structure but involve complex decision-making nonetheless, ranging from choosing a gift to designing a website layout to making a life decision. We hope that our work is a step toward assistants that can help us deliberate and make the best decisions in the range of problems we face every day.

{
\twocolumn[
\begin{center}
\addcontentsline{toc}{section}{References}
\end{center}
]

\section*{Acknowledgments}
We thank Val Ramirez, the data annotators, and the volunteer participants who contributed to our dataset and human evaluation study. We thank the reviewers and action editors for their comments.
The last author thanks Dee Ann Reisinger, Jayant Krishnamurthy, Jason Wolfe, and David Hall for discussing this problem space with him in 2015-2016 and in 2020.

\bibliography{references}
\bibliographystyle{acl_natbib}

}

\clearpage

\appendix

\section{Environment Details}
\label{sec:appendix-env-generation}

Here, we describe how our environments procedurally generate each game, omitting minor details that we implement for task realism. To fully reproduce our environments, please see our code release.

\paragraph{Assignment} To generate a game, each cell of the $k \times k$ table of reviewer-paper affinity scores is sampled from $\text{Uniform}[0,100]$ (with $k=8$ in our experiments).
To ensure that communication is necessary to do well, we reject a random game unless the optimal score with the agents' pooled knowledge is $\geq 1.25$ times as good as the score that either player would achieve with their own information if they replace unknown cells with the average value (50).
For each player independently, we scale the displayed values by a random scalar sampled from $\text{Uniform}[1,10]$.

\paragraph{Planning} To generate contexts for the dialogue, we create a seed list of 39 site names and locations. Each site falls into one of the following categories: restaurants, bars, cafes, sights (museums and landmarks), outdoor (parks), or shopping.

To generate a game, we randomly shuffle the locations of the sites and randomize their features. Each site has five nonzero random features, out of the following list, some of which only apply to some categories: \textit{rating} (categorical), \textit{has parking} (bool), \textit{has takeout} (bool), \textit{touristy} (bool), \textit{cuisine} (categorical), \textit{good for kids} (bool), \textit{accepts reservations} (bool), \textit{open late} (bool), \textit{good for groups} (bool), \textit{ambience} (categorical), \textit{outdoor seating} (bool), \textit{vegetarian options} (bool), \textit{vegan options} (bool), \textit{live music} (bool), \textit{has Wi-Fi} (bool), \textit{alcohol type} (categorical), and \textit{viewpoint} (bool).

We procedurally generate preferences from the user from the following types: 
\vspace{-0.75em}
\begin{itemize}
    \itemsep-0.5em
    \item Feature: a preference over the value of one of the features above
    \item Want to go: a preference to go to a specific site or set of sites
    \item Price: a preference to keep the budget less than some fixed amount
    \item At least one: a preference to go to at least one site of some type (e.g., to visit at least one museum)
    \item Distance: a (negative) preference per unit traveled between sites
\end{itemize}
\vspace{-0.75em}

Each of these preferences is parameterized and randomized on every game. Every user has a price and distance preference; the other preferences are sampled with some probability up to a total of $P$ preferences ($P=10$ in our experiments).
We specifically exclude preference configurations that are counter-intuitive (e.g., a preference for places that do \emph{not} have takeout).
We template natural language descriptions for each preference to present to the user.

\paragraph{Mediation}
To generate a game, we generate a random calendar for each user.  For each 30-min slot between 9am--8pm during a 3-day period, if the slot is still free, we add an event with probability $p_{\text{event}}=0.35$, selecting the event duration uniformly at random from \{30 min, 60 min, 2 hr, 4 hr\}. $f_{\text{shared}}=0.75$ of these events are selected to be shared events that both the assistant and user can see; the remainder are private events that only the user can see. The importance of each event is sampled from $\text{Uniform}[1,10]$.

We generate a set of $F=30$ flights for each user with a random start time in the 3-day period, sampling a duration (in hours) from $\text{Uniform}[1, 10]$. Flight prices for each user $i$ are sampled from $\max(50, \mathcal{N}(\mu_i,\sigma_i))$ to ensure that flight prices a user sees are realistically around the same value, and the parameters of the distribution $\mu=\sigma$ are sampled from $\text{Uniform}[50, 1000]$. We generate a price preference weight $\theta_{\text{price}} \sim \text{Uniform}[-20,-1]$ and preference per 3-hour difference in arrival between the two users' flights $\theta_{\text{arrival}} \sim \text{Uniform}[-10,-1]$ (for every 3 hour difference between their flight times, deduct $\theta_{\text{arrival}}$).

\section{Data Collection Details \& Statistics}
\label{sec:appendix-data-collection}

Human players from Mechanical Turk were vetted via a pre-qualification survey. Data collection was run in multiple dyads, with cooperative players from each dyad (as judged manually) being invited to participate in followup rounds of data collection. Workers are bonused up to \$2.00 in tiers by how close they get to the best possible proposal. In \cref{tab:dataset-stats}, we show the data statistics for human-human dialogues. In \cref{fig:appendix-example-chat-optimization,fig:appendix-example-chat-planning,fig:appendix-example-chat-mediation}, we show example dialogues for each task.

\begin{figure*}[b!]
    \begin{AIbox}{Assignment}
    {\footnotesize
    {\bf A:} I want to make a case for Joseph being Llama \\
    {\bf B:} Hey there! That sounds good to me, my information is limited on LLama. I would like to propose that we assign the following: / Daniel - Electra  / Noah - BLEU / Sofia - SWAG (most confident) / I also see Sofia being solid at some other options too if we should decide we need her to be flexible. \\
    {\bf A:} Okay lets set that up for now  / who do we have for Glove / Ethan is pretty solid \\
    {\bf B:} Lets stick with Ethan for now then. \\
    {\bf A:} he does better at Quac however / unless you have a good Quac option / Ethan can Glove or Quac \\
    {\bf B:} The best person I see for Quac is Noah, but im much more confident having him on BLEU. He is middling at quac / we can put Ethan on Quac \\
    {\bf A:} Ethan Quac selected for now \\
    {\bf B:} Sofia is above average at Glove, its just that I have supreme confidence in her on SWAG. \\
    {\bf A:} okay looks like we dont have much for Glove so we go back / I have no info on Glue for any unselected person \\
    {\bf B:} I dont have any good options for Quac besides going with your information on Ethan, what about you? \\
    {\bf A:} Ethan Quac for now / next best is sofia and she is only slightly worse \\
    {\bf B:} Literally the only info I have for Glue is that Ethan is not so good at it \\
    {\bf A:} Joseph is good at it / only info I have that you dont \\
    {\bf B:} Lets lock in Joseph for now on it, I see Joseph being not so good at anything on my end. / oh wait, you said Joseph on LLama \\
    {\bf A:} yes / thats ideally super lock \\
    {\bf B:} lets go back to that / do you have any good options for swag besides sofia? \\
    {\bf A:} no they are all bad \\
    {\bf B:} trying to see if we might want to move her to roberta since I only see bad options for roberta. / Ok then maybe we will have to take a gamble with roberta / Going to throw a proposal together, feel free to reject once youve absorbed that info. \\
    {\bf A:} do you have any decent options for Andrei Daniel or Morgan / Something mid that isn't taken / I dont have anything like that for them \\
    {\bf B:} Daniel is locked on Electra, but Morgan and Andrei I only see 4/10 levels of confidence. / Do you see anything for Ava? / Andrei could take llama if he really needed to \\
    {\bf A:} for the 3 empty, no / and she is bad at everything taken / Is it a gamble on the last 3 \\
    {\bf B:} do you see anything shes good at thats not taken? Just curious, but most likely we have to gamble with her \\
    {\bf A:} I dont have anything good at them / Yeah the 3 with 0 selection are all red only for me \\
    {\bf B:} Ok here is the gamble version of my proposal haha. \\
    {\bf B:} \texttt{BLEU: Noah Wilson, Electra: Daniel Nguyen, GLUE: Andrei Petrov, GloVe: Ava Li, LLaMA: Joseph Santos, QuAC: Ethan Smith, RoBERTa: Morgan Reed, SWAG: Sofia Patel} \\
    {\bf A:} \texttt{accept}
    }
    \end{AIbox}
    \caption{Example human-human dialogue for \texttt{Assignment}. Forward slashes denote the boundary between multiple messages sent sequentially without a response from the other player.}
    \label{fig:appendix-example-chat-optimization}
\end{figure*}

\begin{figure*}[t!]
    \begin{AIbox}{Planning}
    {\footnotesize
    {\bf User:} Greetings! \\
    {\bf Agent:} Hello! Is there anything in particular that you would like to see in the city? Museums, food options, tourist attractions? \\
    {\bf U:} Yes, I would love to go to a concert at an amazing venue in town. \\
    {\bf A:} I don't have any concerts but there is live music at the Dockside Grill. What is your budget for your trip? \\
    {\bf U:} I'd like to keep it below \$40. \\
    {\bf A:} The Dive has live music for \$30. Does this look like some place you would like to go? \\
    {\bf U:} Yes. \\
    {\bf A:} Ok, we are set on The Dive. With \$10 left in budget, what kind of food would you be interested in? \\
    {\bf U:} Maybe a food truck outside the Mall? \\
    {\bf A:} \texttt{[The Dive, Saul's]} \\
    {\bf A:} With your budget, the nearest food place around \$10 is Sauls but it is a far. There is a closer option with The Dockside Grill but it is \$20 so it will be slightly over budget. Do you prefer either? \\
    {\bf U:} Sauls. And let's find an outdoor attraction I can take kids to, like a park. \\
    {\bf A:} \texttt{[The Dive, Saul's, Garden of Wonders]} \\
    {\bf A:} How does the Garden of Wonders sound? \\
    {\bf U:} \texttt{accept}
    }
    \end{AIbox}
    \vspace{-1mm}
    \caption{Example human-human dialogue for \texttt{Planning}.}
    \label{fig:appendix-example-chat-planning}
    \vspace{-1mm}
\end{figure*}

\newcommand{\roomLmsg}[1]{\parbox{0.45\linewidth}{#1}}
\newcommand{\roomRmsg}[1]{\hspace*{0.51\linewidth}\parbox{0.45\linewidth}{#1}}
\begin{figure*}[b!]
    \centering
    \begin{AIbox}{Mediation}
    \parbox{0.45\linewidth}{{\bf Room 1 (User 1, Assistant)}}\hspace{0.06\linewidth}\parbox{0.45\linewidth}{{\bf Room 2 (User 2, Assistant)}}\\[1ex]
    {\footnotesize
        \roomLmsg{\textbf{U1:} Hello, I've got a busy schedule but need a flight booked. I'm thinking of something Friday evening after 5 PM, but may be able to flex depending on my meetings}\\[1ex]
        \roomLmsg{\textbf{A:} There is a slight overlap but we have a flight on 6/2 4:38 PM — 6:38 PM for 1367\$}\\[1ex]
        \roomRmsg{\textbf{U2:} Hi, I'd like to get a flight, preferably something that doesn't conflict with my important meetings. Is there any redeye flights on Friday, that might work well for me.}\\[1ex]
        \roomRmsg{\textbf{A:} Yes, there is a flight on 6/2 5:28 PM — 2:28 AM that overlaps with your meeting from 2:30 to 6:30 for 629\$}\\[1ex]
        \roomLmsg{\textbf{U1:} That does overlap with a very important meeting, so I wouldn't want to miss any of that. Could we explore other options but I may reconsider if it means I can arrive at the same time as my friend :)}\\[1ex]
        \roomRmsg{\textbf{U2:} That's pretty pricey, maybe I could do a redeye on Wednesday and just miss some of my meeting that evening?}\\[1ex]
        \roomRmsg{\textbf{A:} Would 6/1 6:49 PM — 3:49 AM for 50\$ work? It would overlap with an 8 to 10 PM meeting.}\\[1ex]
        \roomRmsg{\textbf{U2:} That meeting is pretty important, but if it gets me in close to my friend and is that cheap it could be worth it. Go ahead and book it.}\\[1ex]
        \roomLmsg{\textbf{A:} Are you open to 6/1 7:50 PM — 9:50 PM for \$50, it would slightly overlap with the 7:30 PM meeting however it is closer to your friend.}\\[1ex]
        \roomLmsg{\textbf{U1:} I think I may be willing to give up that meeting}\\[1ex]
        \roomLmsg{\textbf{A:} \texttt{Southwest | \$50 | [6/01 07:50 PM] - [6/01 09:50 PM]}}\hspace*{0.06\linewidth}\roomLmsg{\textbf{A:} \texttt{Alaska | \$50 | [6/01 06:49 PM] - [6/02 03:49 AM]}}\\[1ex]
        \roomLmsg{\textbf{U1:} \texttt{accept}}\hspace*{0.06\linewidth}\roomLmsg{\textbf{U2:} \texttt{accept}}
    }
    \end{AIbox}
    \vspace{-1mm}
    \caption{Example human-human dialogue for \texttt{Mediation}.}
    \label{fig:appendix-example-chat-mediation}
    \vspace{-1mm}
\end{figure*}

\begin{table*}[ht!]
\centering
\begin{tabular}{lrrrrr}
\toprule
& Dialogues & Messages ($\mu$) & Words ($\mu$) & Proposals ($\mu$) & Time ($\mu$) \\ \midrule
Assignment & 134 & 18.4 $\pm$ 1.1 & 169.3 $\pm$ 10.9 & 1.7 $\pm$ 0.1 & 8m~9s \\
Planning & 114 & 9.0 $\pm$ 0.4 & 141.9 $\pm$ 6.5 & 3.0 $\pm$ 0.1 & 10m~56s \\
\vspace{0.1cm} Mediation & 162 & 10.9 $\pm$ 0.5 & 119.0 $\pm$ 5.7 & 2.8 $\pm$ 0.2 & 7m~15s \\
\textbf{All Domains} & 409 & 12.8 $\pm$ 0.5 & 141.8 $\pm$ 4.7 & 2.5 $\pm$ 0.1 & 8m~19s \\ 
\bottomrule
\end{tabular}
\caption{Data statistics for human-human dialogues. We collect a total of 409 dialogues, resulting in 5253 messages and 58K words across domains. Dialogues for each setting are roughly the same number of words on average.}\label{tab:dataset-stats}
\end{table*}

\end{document}